\documentclass[runningheads]{llncs}
\usepackage[T1]{fontenc}
\usepackage{xcolor}
\usepackage{graphicx}
\usepackage{amssymb}
\usepackage{amsmath}
\usepackage{bm}
\usepackage{pifont}
\usepackage{multirow}
\usepackage{booktabs}
\usepackage{subfigure}
\usepackage{makecell}
\usepackage{xurl}
\usepackage{array}
\usepackage{xcolor}
\usepackage[export]{adjustbox}
\usepackage{graphicx}
\usepackage[colorlinks,
            linkcolor=blue,
            anchorcolor=blue,
            citecolor=blue]{hyperref}
\usepackage{graphicx,verbatim}

\usepackage{wrapfig}
\usepackage{marvosym}

\definecolor{tabhighlight}{HTML}{e5e5e5}

\begin{document}
\title{Multimodal Large Language Models for Medical Report Generation via Customized Prompt Tuning}
%
\begin{comment}  %% Removed for anonymized MICCAI 2025 submission
\author{First Author\inst{1}\orcidID{0000-1111-2222-3333} \and
Second Author\inst{2,3}\orcidID{1111-2222-3333-4444} \and
Third Author\inst{3}\orcidID{2222--3333-4444-5555}}
%
\authorrunning{F. Author et al.}
% First names are abbreviated in the running head.
% If there are more than two authors, 'et al.' is used.
%
\institute{Princeton University, Princeton NJ 08544, USA \and
Springer Heidelberg, Tiergartenstr. 17, 69121 Heidelberg, Germany
\email{lncs@springer.com}\\
\url{http://www.springer.com/gp/computer-science/lncs} \and
ABC Institute, Rupert-Karls-University Heidelberg, Heidelberg, Germany\\
\email{\{abc,lncs\}@uni-heidelberg.de}}

\end{comment}

\author{Chunlei Li\inst{1}\thanks{Equal contribution.} \and
Jingyang Hou\inst{1}$^\star$ \and
Yilei Shi\inst{1}$^\star$ \and
Jingliang Hu\inst{1} \and
Xiao Xiang Zhu\inst{2} \and
Lichao Mou\inst{1}\textsuperscript{(\Letter)}}

\authorrunning{C. Li et al.}
% First names are abbreviated in the running head.
% If there are more than two authors, 'et al.' is used.
%
\institute{MedAI Technology (Wuxi) Co. Ltd., Wuxi, China\\\email{lichao.mou@medimagingai.com} \and Technical University of Munich, Munich, Germany}

\maketitle              % typeset the header of the contribution
\begin{abstract}
Medical report generation from imaging data remains a challenging task in clinical practice. While large language models (LLMs) show great promise in addressing this challenge, their effective integration with medical imaging data still deserves in-depth exploration. In this paper, we present MRG-LLM, a novel multimodal large language model (MLLM) that combines a frozen LLM with a learnable visual encoder and introduces a dynamic prompt customization mechanism. Our key innovation lies in generating instance-specific prompts tailored to individual medical images through conditional affine transformations derived from visual features. We propose two implementations: prompt-wise and promptbook-wise customization, enabling precise and targeted report generation. Extensive experiments on IU X-ray and MIMIC-CXR datasets demonstrate that MRG-LLM achieves state-of-the-art performance in medical report generation. Our code will be made publicly available.

% Authors must provide keywords and are not allowed to remove this Keyword section.

\end{abstract}
\section{Introduction}

\label{sec:intro}
Medical report writing is a critical component of clinical practice, serving as the primary means of documenting patient conditions, communicating diagnostic findings, and planning treatment strategies. The process of generating these reports, however, is often time-consuming and labor-intensive, placing a significant burden on healthcare professionals. This challenge has motivated extensive research into automatic medical report generation methods, which aim to assist clinicians by automatically producing accurate and detailed reports from medical images, ultimately improving healthcare efficiency and reducing physician workload.
\par
Recent years have witnessed substantial progress in automatic medical report generation, with numerous approaches proposed to address this challenging multi-modal task. Prior methods predominantly adopt an encoder-decoder paradigm and are trained from scratch to transform images into textual descriptions. While these approaches have shown promise, they face inherent limitations in their text generation capabilities, primarily due to the restricted volume of available training data in the medical domain. This scarcity stems from privacy concerns and the sensitive nature of medical records. Unlike general image captioning tasks in computer vision where large-scale datasets are readily available from the internet, medical report generation must contend with relatively small datasets, making it challenging for training sophisticated text decoders from scratch. For instance, widely used medical datasets such as IU X-ray and MIMIC-CXR contain only 4K and 220K images, respectively, which are significantly smaller compared to general image captioning datasets like Conceptual Captions (3.3M images) and Conceptual 12M (12M images).
\par
The emergence of large language models (LLMs) presents a compelling opportunity to address these limitations. Recent advances in LLMs, particularly decoder-only architectures like GPT~\cite{ref_gpt} and Llama~\cite{ref_llama}, have demonstrated remarkable capabilities in natural language understanding and generation across diverse domains. These models, pre-trained on vast amounts of text data, possess rich knowledge representations and sophisticated reasoning abilities that can potentially enhance medical report generation. Their decoder-only architecture, combined with extensive pre-training, enables them to capture complex linguistic patterns and generate more natural and contextually appropriate texts compared to traditional approaches.
\par
Several pioneering works have begun exploring the integration of LLMs into medical report generation frameworks. Notable examples such as R2GenGPT~\cite{ref_r2gengpt} and RaDialog~\cite{ref_radialog} demonstrate the feasibility of leveraging frozen LLMs in combination with learnable visual encoders. These approaches typically employ carefully crafted language instructions to guide LLMs in generating medical reports, achieving promising results through cross-modal alignment between visual and textual representations. However, manually designing optimal instructions for diverse medical scenarios poses significant challenges, as it requires extensive domain expertise and clinical knowledge.
\par
In this work, we leverage prompt tuning with learnable prompts. While prompt tuning has demonstrated effectiveness in various natural language processing tasks, we observe that generic prompts fail to adequately capture the unique characteristics present in individual medical images. Medical images exhibit diverse pathological findings, anatomical structures, and visual patterns that require specialized attention and description. This observation motivates our key innovation: customized prompt tuning for medical report generation. Rather than relying on general prompts, we introduce a dynamic prompt customization mechanism that generates instance-specific prompts tailored to each medical image. Our approach learns to manipulate prompts through a conditional affine transformation based on visual features of the input images. This personalization enables more precise and targeted report generation by allowing the model to adapt its prompting strategy based on the specific visual content under examination.
\par
Our main contributions can be summarized as follows:
\begin{itemize}
\item We introduce MRG-LLM, a multimodal large language model (MLLM) for medical report generation that combines a frozen LLM with a learnable visual encoder and customized prompt tuning, enabling more accurate and personalized report generation.
\item We propose a novel prompt customization mechanism that dynamically adapts soft prompts based on visual features, implementing this through two specific instantiations: prompt-wise and promptbook-wise customization.
\item Through extensive experiments on two public datasets, we demonstrate that our approach achieves state-of-the-art performance, outperforming existing methods.
\end{itemize}

\section{Methodology}
\label{sec:method}
This section delves into the architecture of MRG-LLM, which generates reports from medical images using customized prompts. We first describe MRG-LLM's core components, followed by the proposed prompt customization mechanism and training strategy.

\begin{figure}
    \centering
    \includegraphics[width=\textwidth]{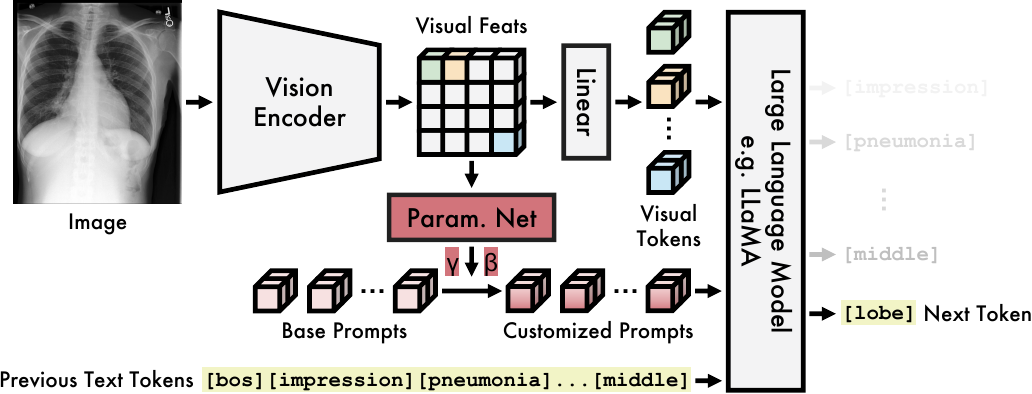}
    \caption{Overview of our MRG-LLM framework for medical report generation.}
    \label{fig:main_image}
\end{figure}

\subsection{Model Architecture}

As illustrated in Fig.~\ref{fig:main_image}, MRG-LLM consists of three primary components: a vision encoder $f_{ve}$, a projection layer $f_{proj}$, and a pre-trained LLM backbone $f_{llm}$. This architecture enables effective cross-modal learning between visual and textual domains while maintaining the powerful language generation capability of the LLM.
\par
The vision encoder $f_{ve}$ takes as input an image $\mathbf{I} \in \mathbb{R}^{H \times W \times C}$, where $H$ and $W$ represent the height and width of the image, and $C$ denotes the number of channels. The encoder outputs a sequence of visual features $\mathbf{X} = (\mathbf{x}_1, \mathbf{x}_2, \ldots, \mathbf{x}_M)$
$\in \mathbb{R}^{M \times C^{\prime}}$, where $M$ is the number of features, and $C^{\prime}$ is the feature dimension. The projection $f_{proj}$ is a linear layer that maps the visual features $\mathbf{X}$ to visual tokens $\mathbf{V} = (\mathbf{v}_1, \mathbf{v}_2, \ldots, \mathbf{v}_M) \in \mathbb{R}^{M \times D}$, where $D$ matches the dimension of the LLM's embedding space.
\par
The LLM backbone $f_{llm}$, based on a decoder-only Transformer architecture, processes a mixed sequence of tokens $\mathbf{Z} = (\mathbf{z}_1, \mathbf{z}_2, \ldots, \mathbf{z}_K) \in \mathbb{R}^{K \times D}$ comprising visual tokens, text tokens, and learnable prompt tokens, where $K$ represents the total number of tokens. Through self-attention mechanism, the LLM backbone captures contextual relationships between different token types.
\par
The output of the LLM is a sequence of predicted tokens $\mathbf{T} = (\mathbf{t}_1, \mathbf{t}_2, \ldots, \mathbf{t}_K)$
$\in \mathbb{R}^{K \times D}$. Each token prediction $\mathbf{t}_i$ is conditioned on all previous tokens $\mathbf{Z}_{<i} = (\mathbf{z}_1, \ldots, \mathbf{z}_{i-1})$:
\begin{equation}
\mathbf{t}_i = f_{llm}(\mathbf{Z}_{<i}) \,.
\end{equation}
\par
Finally, $\mathbf{t}_i$ is passed through a final linear layer followed by a softmax, mapping hidden states to vocabulary probabilities. This layer is denoted as $f_{vocab}: \mathbb{R}^{D} \to \mathbb{R}^{V}$, where $V$ is the size of the vocabulary. The final prediction $\tilde{\mathbf{w}}_i$ is determined by selecting the vocabulary word with the highest probability:
\begin{equation}
\tilde{\mathbf{w}}_i = \arg\max_{k\in \text{vocab}} f_{vocab}(\mathbf{t}_i)[k] \,.
\end{equation}

\subsection{Prompt Customization}

While existing LLM-based approaches typically employ task-wise prompting with shared prompts across all medical images~\cite{ref_r2gengpt,ref_radialog}, we recognize the need for personalized prompt tuning in medical report generation. We propose instance-wise prompting, which dynamically adapts prompts for each radiographic study to generate more precise reports.
\par
We first define a \textbf{\textit{promptbook}} $\mathbf{P} \in \mathbb{R}^{N \times D}$, comprising $N$ learnable base prompts of dimension $D$. Our prompt customization module learns to modify these base prompts through an affine transformation conditioned on the visual content of each input image. We propose two approaches for implementing this idea.

\vspace{0.5\baselineskip}
\noindent
\textbf{Prompt-Wise Customization} We learn a function $\phi$ that generates image-specific transformation parameters:
\begin{equation}
(\boldsymbol{\gamma}, \boldsymbol{\beta}) = \phi(\mathbf{X}) \,,
\end{equation}
where $\boldsymbol{\gamma} \in \mathbb{R}^{N}$ and $\boldsymbol{\beta} \in \mathbb{R}^{N}$ modulate each base prompt $\mathbf{p}_i$, whose subscript refers to the $i$-th learnable prompt in the promptbook $\mathbf{P}$, via a prompt-wise affine transformation:
\begin{equation}
\mathbf{p}_i^\prime = \gamma_i\mathbf{p}_i + \beta_i \,.
\end{equation}

\vspace{0.5\baselineskip}
\noindent
\textbf{Promptbook-Wise Customization} This instantiation uses a global transformation to the entire promptbook:
\begin{equation}
(\gamma, \beta) = \phi(\mathbf{X}) \,,
\end{equation}
where $\gamma$ and $\beta$ are scalars that scale and shift the entire promptbook $\mathbf{P}$:
\begin{equation}
\mathbf{P}^\prime = \gamma\mathbf{P} + \beta \,.
\end{equation}
\par
This streamlined approach requires only two parameters, making it computationally efficient while maintaining effective prompt customization.

\subsection{Training and Inference}
Our framework implements an autoregressive pipeline for both training and inference. We append \verb|[BOS]| and \verb|[EOS]| tokens to all reports.
\par
During training, our model takes a concatenation of visual tokens, learnable prompts, and previously predicted tokens as input to predict the next token. We optimize the model by minimizing the cross-entropy loss between the predicted report and the corresponding ground truth report:
\begin{equation}
\mathcal{L}_{\text{ce}} = - \sum_{i=1}^t \mathbf{w}_i\log\tilde{\mathbf{w}}_i \,.
\end{equation}
\par
For inference, the model generates texts autoregressively. Given a test image, we first provide a \verb|[BOS]| token alongside the image's visual tokens and learnable prompts to predict the first token. This predicted token is then concatenated with the previous input to form a new input sequence for the LLM backbone, which predicts the second token. The process continues iteratively, using the previously predicted tokens to predict each subsequent token until an \verb|[EOS]| token is generated, signaling the end of generation.

\section{Experiments}
\label{sec:experiments}
\subsection{Datasets}
\label{sec:datasets}
We evaluate our approach on two widely used datasets. MIMIC-CXR~\cite{ref_mimic} consists of 377,110 chest X-ray images paired with 227,835 radiology reports. Following the official splits, we use 368,960 images and 222,758 reports for training, 2,991 images and 1,808 reports for validation, and 5,159 images and 3,269 reports for testing. IU X-ray~\cite{ref_iuxray} contains 7,470 chest X-ray images with 3,955 associated radiology reports. Following prior work~\cite{ref_iuxray}, we employ a 7:1:2 patient-disjoint split for training, validation, and testing.

\subsection{Experimental Settings}
\label{sec:settings}

\vspace{0.5\baselineskip}
\noindent
\textbf{Metrics}
We adopt three standard metrics for evaluation: BLEU, ROUGE-L, and METEOR.

\vspace{0.5\baselineskip}
\noindent
\textbf{Implementation Details}
We extract visual features using a ConvNeXt-Tiny backbone pretrained on ImageNet-1K. The backbone outputs 7$\times$7 feature maps with 768 channels, which are projected to 4096 dimensions (i.e., $M=49$, $C^{\prime}=768$, $D=4096$). We employ 50 learnable prompts and train the model using Adam optimizer with a batch size of 6 and learning rate of 1e-4. Training runs for 15 epochs on IU X-ray and 10 epochs on MIMIC-CXR. All experiments are conducted using PyTorch on 4 NVIDIA GeForce RTX 4090 GPUs.

\subsection{Comparison with State-of-the-Art Methods}

\begin{table}[t]
\setlength{\abovecaptionskip}{2pt}
\caption{Quantitative comparison with state-of-the-art methods on IU X-ray and MIMIC-CXR datasets. $\ast$ denotes results reproduced using official implementations. $\dag$ indicates results without dataset-specific pre-training. \textbf{Bold} and \underline{underlined} values indicate the best and second-best performance, respectively.}
\renewcommand{\arraystretch}{1.0}
\centering
\resizebox{1.0\linewidth}{!}{
\begin{tabular}{l|>{\hspace{0.1cm}}cccccc<{\hspace{0.1cm}}| >{\hspace{0.1cm}} cccccc}
\toprule
& \multicolumn{6}{c|}{\textbf{IU X-ray}} & \multicolumn{6}{c}{\textbf{MIMIC-CXR}} \\ 
& \textbf{BL-1} & \textbf{BL-2} & \textbf{BL-3} & \textbf{BL-4} & \textbf{RG-L} & \textbf{MTR} & \textbf{BL-1} & \textbf{BL-2} & \textbf{BL-3} & \textbf{BL-4} & \textbf{RG-L} & \textbf{MTR} \\ 
\midrule
Show-Tell~\cite{ref_show_tell} & 0.243 & 0.130 & 0.108 & 0.078 & 0.307 & 0.157 & 0.308 & 0.190 & 0.125 & 0.088 & 0.256 & 0.122 \\
Transformer~\cite{ref_transformer} & 0.372 & 0.251 & 0.147 & 0.136 & 0.317 & 0.168 & 0.316 & 0.199 & 0.140 & 0.092 & 0.267 & 0.129 \\
Att2in~\cite{ref_att2in} & 0.248 & 0.134 & 0.116 & 0.091 & 0.309 & 0.162 & 0.314 & 0.198 & 0.133 & 0.095 & 0.264 & 0.122 \\
AdaAtt~\cite{ref_adaatt} & 0.284 & 0.207 & 0.150 & 0.126 & 0.311 & 0.165 & 0.314 & 0.198 & 0.132 & 0.094 & 0.267 & 0.128 \\
Up-Down~\cite{ref_updown} & - & - & - & - & - & - & 0.317 & 0.195 & 0.130 & 0.092 & 0.267 & 0.128 \\
$\mathcal{M}^{2}$ Transf.~\cite{ref_m2transformer} & 0.402 & 0.284 & 0.168 & 0.143 & 0.328 & 0.170 & 0.332 & 0.210 & 0.142 & 0.101 & 0.264 & 0.134\\
R2Gen~\cite{ref_r2gen} & 0.470 & 0.304 & 0.219 & 0.165 & 0.371 & 0.187 & 0.353 & 0.218 & 0.145 & 0.103 & 0.277 & 0.142 \\
SentSAT+KG~\cite{ref_sentsat_kg} & 0.441 & 0.291 & 0.203 & 0.147 & 0.367 & - & - & - & - & - & - & - \\
PPKED~\cite{ref_ppked} & 0.483 & 0.315 & 0.224 & 0.168 & 0.376 & - & 0.360 & 0.224 & 0.149 & 0.106 & 0.284 & 0.149 \\
Contr. Attn.~\cite{ref_co_attention} & 0.492 & 0.314 & 0.222 & 0.169 & 0.381 & 0.193 & 0.350 & 0.219 & 0.152 & 0.109 & 0.283 & 0.151 \\
CMCL~\cite{ref_cmcl} & 0.473 & 0.305 & 0.217 & 0.162 & 0.378 & 0.186 & 0.344 & 0.217 & 0.140 & 0.097 & 0.281 & 0.133 \\
CMN~\cite{ref_cmn} & 0.475 & 0.309 & 0.222 & 0.170 & 0.375 & 0.191 & 0.353 & 0.218 & 0.148 & 0.106 & 0.278 & 0.142 \\
AlignTransf.~\cite{ref_aligntransformer} & 0.484 & 0.313 & 0.225 & 0.173 & 0.379 & \underline{0.204} & 0.378 & 0.235 & 0.156 & 0.112 & 0.283 & 0.158 \\
$\mathcal{M}^{2}$ \textsc{Tr}. \textsc{Prog.}~\cite{ref_m2tr} & 0.486 & 0.317 & 0.232 & 0.173 & 0.390 & 0.192 & 0.378 & 0.232 & 0.154 & 0.107 & 0.272 & 0.145 \\
CMM+RL~\cite{ref_cmm_rl} & 0.494 & 0.321 & \underline{0.235} & \underline{0.181} & 0.384 & 0.201 & 0.381 & 0.232 & 0.155 & 0.109 & \underline{0.287} & 0.151 \\
XPRONET$^\ast$~\cite{ref_xpronet} & 0.491 & \underline{0.325} & 0.228 & 0.169 & 0.387 & 0.202 &0.344 & 0.215 & 0.146 & 0.105 & 0.279 & 0.138 \\
MCGN~\cite{ref_mcgn} & 0.481 & 0.316 & 0.226 & 0.171 & 0.372 & 0.190 & 0.373 & 0.235 & 0.162 & 0.120 & 0.282 & 0.143 \\
M2KT~\cite{ref_m2kt} & \underline{0.497} & 0.319 & 0.230 & 0.174 & \underline{0.399} & - & 0.386 & 0.237 & 0.157 & 0.111 & 0.274 & - \\
RAMT~\cite{ref_ramt} & 0.482 & 0.310 & 0.221 & 0.165 & 0.377 & 0.195 & 0.362 & 0.229 & 0.157 & 0.113 & 0.284 & 0.153 \\
R2GenGPT~\cite{ref_r2gengpt} & 0.482 & 0.306 & 0.215 & 0.158 & 0.370 & 0.200 & 0.387 & 0.248 & \underline{0.170} & 0.123 & 0.280 & 0.149 \\
VLCI$^\dag$~\cite{ref_vlci} & 0.324 & 0.211 & 0.151 & 0.115 & 0.379 & 0.166 & 0.357 & 0.216 & 0.144 & 0.103 & 0.256 & 0.136 \\
RaDialog-RG~\cite{ref_radialog} & - & - & - & - & - & - & 0.346 & - & - & 0.095 & 0.271 & 0.140 \\
PromptMRG~\cite{ref_promptmrg} & 0.401 & - & - & 0.098 & 0.281 & 0.160 & \underline{0.398} & - & - & 0.112 & 0.268 & 0.157 \\
AdaMatch-Cyclic~\cite{ref_adamatch} & - & - & - & - & - & - & 0.379 & 0.235 & 0.154 & 0.106 & 0.286 & \underline{0.163} \\
HERGen~\cite{ref_hergen} & - & - & - & - & - & - & 0.395 & \underline{0.248} & 0.169 & \underline{0.122} & 0.285 & 0.156 \\
MedRAT~\cite{ref_medrat} & 0.455 & - & - & 0.129 & 0.349 & - & 0.365 & - & - & 0.086 & 0.251 & - \\
\textbf{Ours} & \textbf{0.529} & \textbf{0.359} & \textbf{0.266} & \textbf{0.202} & \textbf{0.408} & \textbf{0.221} & \textbf{0.416} & \textbf{0.267} & \textbf{0.182} & \textbf{0.129} & \textbf{0.296} & \textbf{0.163} \\
\bottomrule
\end{tabular}
}
\label{tab:comparision_of_sota}
\end{table}

\begin{figure}[h!]
    \centering
    \includegraphics[width=\textwidth]{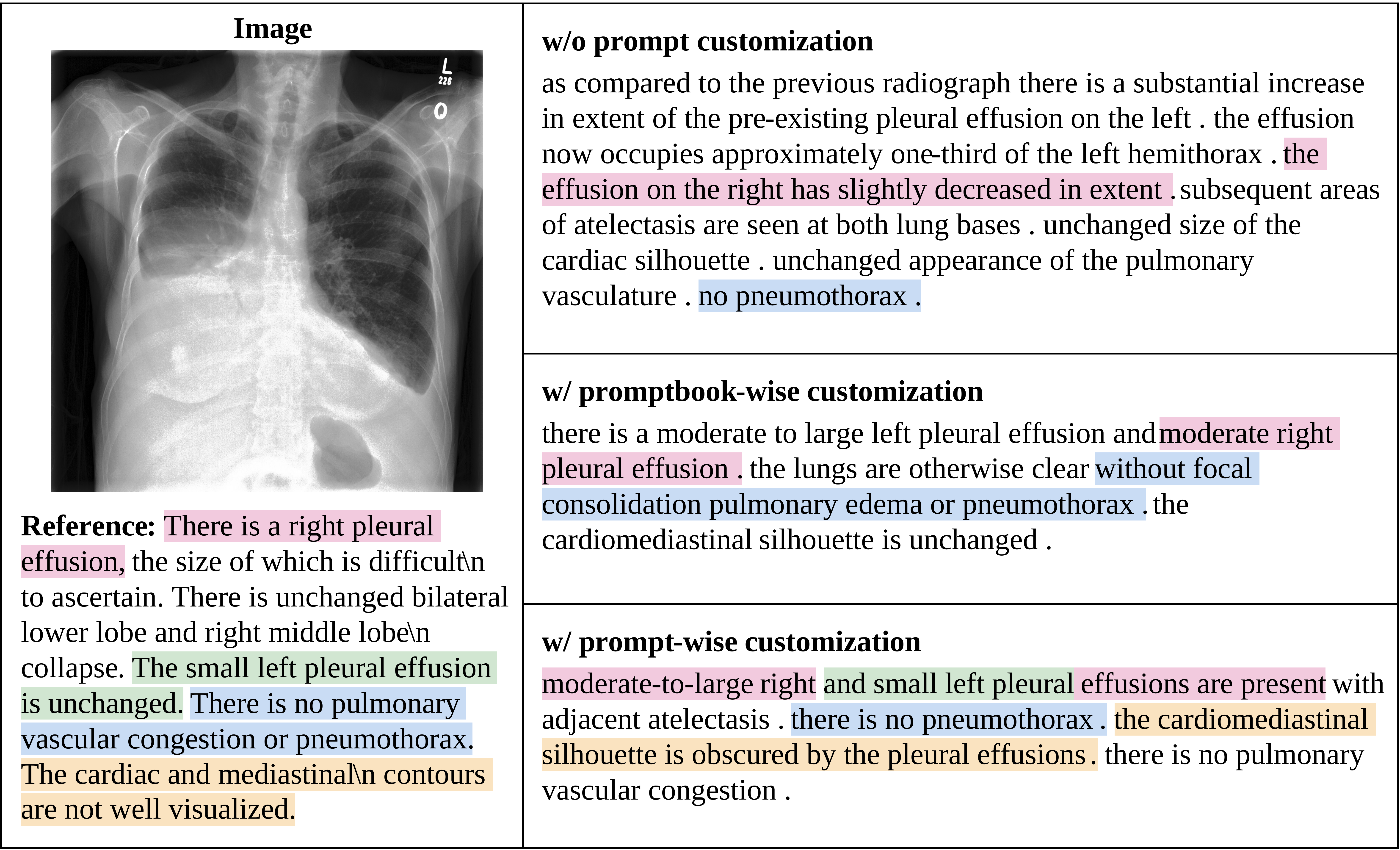}
    \caption{Visualization of a sample with reports from the MIMIC-CXR dataset. The baseline employs learnable prompts without customization, while our two proposed prompt customization approaches demonstrate improved performance. Text with the same color highlights descriptions of the same disease across different methods.}
    \label{fig:combined_plots}
\end{figure}

We benchmark MRG-LLM against 27 baseline methods, including 6 general image captioning approaches (Show-Tell~\cite{ref_show_tell}, Transformer~\cite{ref_transformer}, Att2in~\cite{ref_att2in}, AdaAtt~\cite{ref_adaatt}, UP-DOWN~\cite{ref_updown}, and $\mathcal{M}^{2}$ Transformer~\cite{ref_m2transformer}) and 21 specialized medical report generation models, including recent works like HERGen~\cite{ref_hergen} and MedRAT~\cite{ref_medrat}. Table~\ref{tab:comparision_of_sota} present the comprehensive results.
\par
As shown in Table~\ref{tab:comparision_of_sota}, MRG-LLM achieves state-of-the-art performance on IU X-ray, with absolute improvements of 6.4\%, 11.6\%, and 8.3\% in BLEU-1, BLEU-4, and METEOR, respectively. On MIMIC-CXR, our method outperforms previous approaches across all metrics. 

\subsection{Discussion}

\vspace{0.5\baselineskip}
\noindent
\textbf{Prompt Customization} 
Our experiments demonstrate that customized prompt tuning outperforms traditional task-wise prompt tuning (cf. Table~\ref{tab:learnable_prompts_comparison} and Fig.~\ref{fig:combined_plots}). Between our two instantiations, prompt-wise customization achieves superior results, improving BLEU-1 from 0.395 to 0.416, ROUGE-L from 0.289 to 0.296, and CIDEr from 0.224 to 0.258. Notably, promptbook-wise customization yields significant gains despite using only two parameters.

\begin{table}[t]
    \renewcommand{\arraystretch}{1.0}
    \centering
    \caption{Quantitative comparison of different prompting strategies.} 
    \resizebox{0.8\linewidth}{!}{ % 调整表格宽度到页面宽度
    \begin{tabular}{r<{\hspace{0.1cm}}|>{\hspace{0.1cm}}cccccc}
    \midrule[0.85pt]
     & \textbf{BL-1} & \textbf{BL-2} & \textbf{BL-3} & \textbf{BL-4} & \textbf{RG-L} & \textbf{MTR} \\ 
    \midrule
    w/o prompt customization & 0.395 & 0.252 & 0.172 & 0.123 & 0.289 & 0.155 \\
    w/ prompt-wise customization & \textbf{0.416} & \textbf{0.267} & \textbf{0.182} & \textbf{0.129} & \textbf{0.296} & \textbf{0.163} \\
    w/ promptbook-wise customization & 0.413 & 0.264 & 0.180 & 0.127 & 0.293 & 0.162 \\
    \midrule[0.85pt]
    \end{tabular}}
    \label{tab:learnable_prompts_comparison}
\end{table}

\vspace{0.5\baselineskip}
\noindent
\textbf{Analysis of Transformation Parameters}
We analyze the individual contributions of $\boldsymbol{\gamma}$ and $\boldsymbol{\beta}$ through ablation studies (see Table~\ref{tab:effect_gamma_beta}). Training with only $\boldsymbol{\beta}$ or $\boldsymbol{\gamma}$ results in performance drops of 2.9\% and 0.1\% respectively, indicating $\boldsymbol{\gamma}$’s greater importance. This finding is further reinforced by test-time ablations: removing $\boldsymbol{\beta}$ causes a 4.2\% decrease, while removing $\boldsymbol{\gamma}$ leads to a substantial 60.3\% drop in performance.

\begin{table}[t]
\renewcommand{\arraystretch}{1.0}
\centering
\caption{Ablation study analyzing the impact of $\boldsymbol{\gamma}$ and $\boldsymbol{\beta}$ in various training and inference configurations. Results demonstrate performance differences with and without them.}
\resizebox{0.60\linewidth}{!}{
\begin{tabular}{r!{\hspace{0.1cm}}|!{\hspace{0.2cm}}c!{\hspace{0.2cm}}!{\hspace{0.2cm}}c!{\hspace{0.2cm}}|!{\hspace{0.1cm}}cccccc} 
\toprule
 & $\boldsymbol{\gamma}$  & $\boldsymbol{\beta}$  & \textbf{BL-1} & \textbf{BL-2} & \textbf{BL-3} & \textbf{BL-4} & \textbf{RG-L} & \textbf{MTR} \\ 
\midrule
\multirow{3}{*}{Training} & \checkmark & \checkmark  & 0.416  & 0.267  & 0.182  & 0.129  & 0.296  &  0.163 \\
 & \checkmark & -  & 0.409  & 0.267  & 0.185  & 0.133  & 0.298  &  0.161 \\ 
 &  - & \checkmark  & 0.398  & 0.258  & 0.178  & 0.129  & 0.293  &  0.157 \\
\midrule
\multirow{2}{*}{Inference} & \checkmark & - & 0.402  & 0.257  & 0.175  & 0.124  & 0.292  &  0.157  \\ 
 &  -  & \checkmark  & 0.240  & 0.109  & 0.043  & 0.016  & 0.158  &  0.087 \\ 
\bottomrule
\end{tabular}}
%\vspace{-0.4cm}
\label{tab:effect_gamma_beta}
\end{table}

\vspace{0.5\baselineskip}
\noindent
\textbf{Parameter Network Architecture}
We investigate the impact of parameter network $\phi$ depth on transformation parameter learning. As shown in Table~\ref{table:para_comparison}, a 2-layer MLP achieves optimal performance compared to both simpler (linear layer) and more complex (3-layer MLP) architectures, motivating our final choice.

\begin{table*}[h]
    \renewcommand{\arraystretch}{1.0}
    \centering
    \caption{Ablation study examining the influence of parameter net depth on model performance.} 
    \label{table:para_comparison}
    \resizebox{0.56\linewidth}{!}{
    \begin{tabular}{r<{\hspace{0.1cm}}|>{\hspace{0.1cm}}ccccccc}
    \midrule[0.85pt]
     & \textbf{BL-1} & \textbf{BL-2} & \textbf{BL-3} & \textbf{BL-4} & \textbf{RG-L} & \textbf{MTR} \\ 
    \midrule
    Linear & 0.406 & 0.261 & 0.178 & 0.127 & 0.291 & 0.159 \\
    2-layer MLP & \textbf{0.416} & \textbf{0.267} & \textbf{0.182} & \textbf{0.129} & \textbf{0.296} & \textbf{0.163} \\
    3-layer MLP & 0.411 & 0.263 & 0.178 & 0.125 & 0.294 & 0.162 \\
    \midrule[0.85pt]
    \end{tabular}}
\end{table*}

\begin{wrapfigure}{r}{6cm}
\vspace{-1.0cm}
 \centering
 \includegraphics[width=0.42\textwidth]{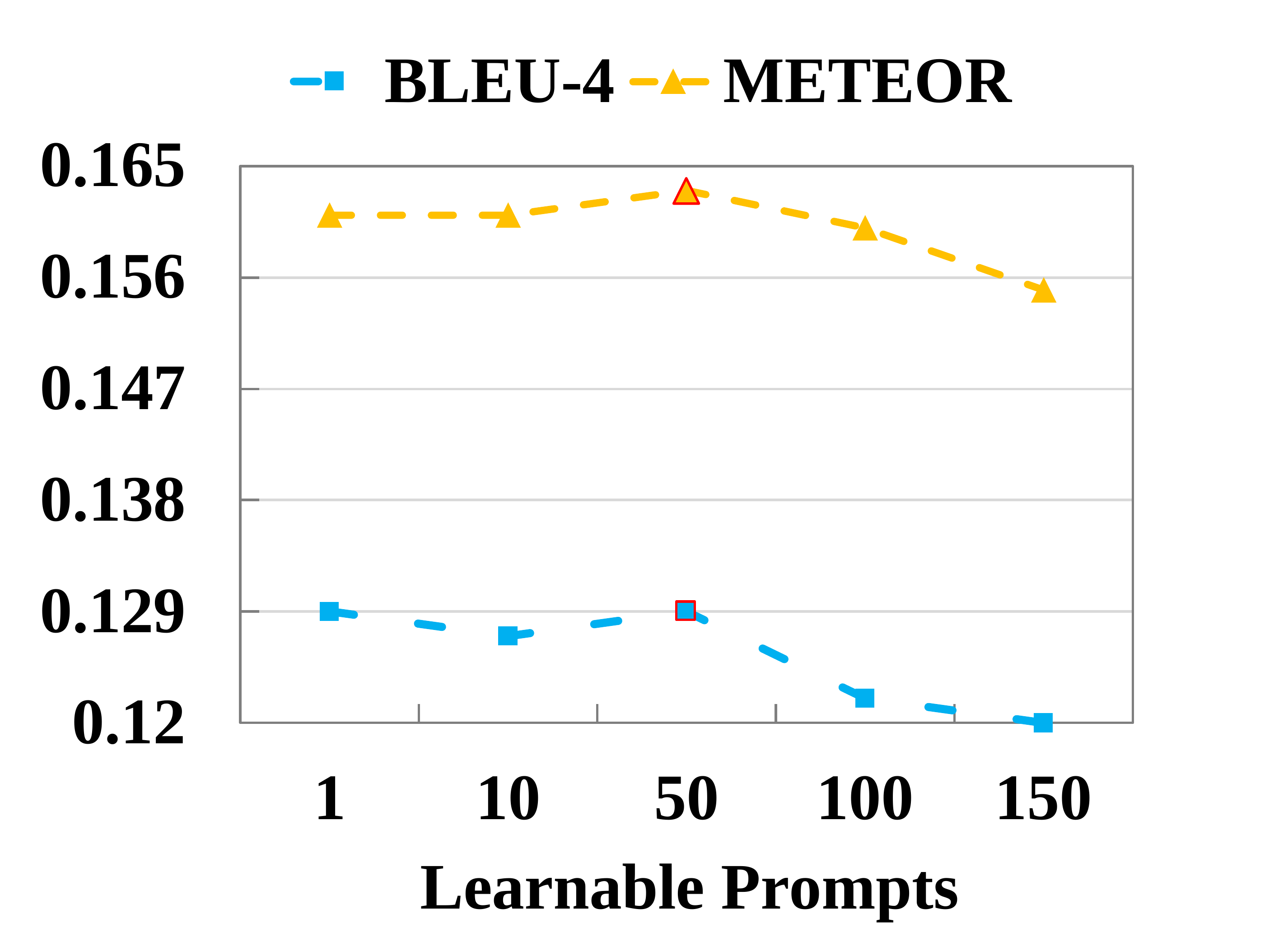}
 \caption{Ablation study on the impact of varying the number of learnable prompts.}
 \label{fig:learnable_prompts}
 \vspace{-1.0cm}
\end{wrapfigure}

\vspace{0.5\baselineskip}
\noindent
\textbf{Number of Learnable Prompts} We investigate how the number of learnable prompts affects MRG-LLM's performance on the MIMIC-CXR dataset (cf. Fig.~\ref{fig:learnable_prompts}). Our experiments reveal that optimal performance is achieved when the number of learnable prompts approaches the number of visual tokens. Notably, even a single learnable prompt demonstrates competitive performance.

\section{Conclusion}
We present MRG-LLM, a framework for medical report generation that leverages LLMs. By combining a frozen LLM with a learnable visual encoder and introducing instance-specific prompt customization, our approach enables more accurate medical report generation. Experiments on IU X-ray and MIMIC-CXR datasets demonstrate that our dynamic prompt customization mechanism achieves state-of-the-art performance. Future work could explore more sophisticated prompt customization methods and extension to diverse medical imaging modalities. 

\begin{comment}  %% removed for anonymized MICCAI 2025 submission.
    
    % The following acknowledgement and disclaimer sections should be removed for the double-blind review process.  
    % If and when your paper is accepted, reinsert the acknowledgement and the disclaimer clause in your final camera-ready version.

\begin{credits}
\subsubsection{\ackname} A bold run-in heading in small font size at the end of the paper is
used for general acknowledgments, for example: This study was funded
by X (grant number Y).

\subsubsection{\discintname}
It is now necessary to declare any competing interests or to specifically
state that the authors have no competing interests. Please place the
statement with a bold run-in heading in small font size beneath the
(optional) acknowledgments\footnote{If EquinOCS, our proceedings submission
system, is used, then the disclaimer can be provided directly in the system.},
for example: The authors have no competing interests to declare that are
relevant to the content of this article. Or: Author A has received research
grants from Company W. Author B has received a speaker honorarium from
Company X and owns stock in Company Y. Author C is a member of committee Z.
\end{credits}

\end{comment}
%
% ---- Bibliography ----
%
% BibTeX users should specify bibliography style 'splncs04'.
% References will then be sorted and formatted in the correct style.
%
% \bibliographystyle{splncs04}
% \bibliography{mybibliography}
%

\end{document}